\documentclass{article}
\input{setting.tex}

\usepackage{fullpage}

\title{\mname: Graph-based Attention Model for Healthcare Representation Learning}

\author{
Edward Choi$^*$, Mohammad Taha Bahadori$^*$, Le Song$^*$, Walter F. Stewart$^\dagger$, Jimeng Sun$^*$\\
$^*$ Georgia Institute of Technology \qquad $^\dagger$ Sutter Health \\ 
\small \texttt{\{mp2893,bahadori\}@gatech.edu}, \texttt{lsong@cc.gatech.edu}, \texttt{stewarwf@sutterhealth.org}, \texttt{jsun@cc.gatech.edu} \normalsize
}

\date{}
\begin{document}

\maketitle

\begin{abstract}
Deep learning methods exhibit promising performance for predictive modeling in healthcare, but two important challenges remain:
\begin{itemize}[leftmargin=5.5mm]
\item {\it Data insufficiency:} Often in healthcare predictive modeling, the sample size is insufficient for deep learning methods to achieve satisfactory results. 
\item {\it Interpretation:} The representations learned by deep learning methods should align with medical knowledge.
\end{itemize}
To address these challenges, we propose a GRaph-based Attention Model, \mname that supplements electronic health records (EHR) with hierarchical information inherent to medical ontologies. 
Based on the data volume and the ontology structure, \mname represents a medical concept as a combination of its ancestors in the ontology via an attention mechanism. 

We compared predictive performance (\textit{i.e.} accuracy, data needs, interpretability) of \mname to various methods including the recurrent neural network (RNN) in two sequential diagnoses prediction tasks and one heart failure prediction task.
Compared to the basic RNN, \mname achieved 10\% higher accuracy for predicting diseases rarely observed in the training data and 3\% improved area under the ROC curve for predicting heart failure using an order of magnitude less training data. Additionally, unlike other methods, the medical concept representations learned by \mname are well aligned with the medical ontology. Finally, \mname exhibits intuitive attention behaviors by adaptively generalizing to higher level concepts when facing data insufficiency at the lower level concepts.
\end{abstract}

%\vspace*{-3mm}
\section{Introduction}
\label{sec:intro}
The rapid growth in volume and diversity of health care data from electronic health records (EHR) and other sources is motivating the use of predictive modeling to improve care for individual patients. In particular, novel applications are emerging that use deep learning methods such as word embedding \citep{choi2016multi,choi2016learning}, recurrent neural networks (RNN) \citep{che2016recurrent,choi2016doctor,choi2016retain,lipton2016modeling}, convolutional neural networks (CNN) \citep{nguyen2016deepr} or stacked denoising autoencoders (SDA) \citep{che2015deep,miotto2016deep}, demonstrating significant performance enhancement for diverse prediction tasks. Deep learning models appear to perform significantly better than logistic regression or multilayer perceptron (MLP) models that depend, to some degree, on expert feature construction \citep{lipton2015learning,razavian2016multi}.

Training deep learning models typically requires large amounts of data that often cannot be met by a single health system or provider organization. Sub-optimal model performance can be particularly challenging when the focus of interest is predicting onset of a rare disease. For example, using Doctor AI \citep{choi2016doctor}, we discovered that RNN alone was ineffective to predict the onset of diseases such as cerebral degenerations (e.g. Leukodystrophy, Cerebral lipidoses) or developmental disorders (e.g. autistic disorder, Heller’s syndrome), partly because their rare occurrence in the training data provided little learning opportunity to the flexible models like RNN.

%We typically need large amount of data. This data requirement is often unmet in healthcare applications because EHR data from individual provider organizations are often limited. And cohort construction, a common process in HPM to create a set of patients for a specific study (\textit{e.g.} heart failure onset prediction), typically decreases the size of the dataset because of various exclusion criteria. Also, when it comes to less common diseases the data itself is scarce in all healthcare organizations, which can lead to poor predictive performance. For example, using Doctor AI \citep{choi2016doctor}, we discovered that RNN alone was ineffective for less common diseases such as cerebral degenerations (\textit{e.g.} Leukodystrophy, Cerebral lipidoses) and developmental disorders (\textit{e.g.} autistic disorder, Heller's syndrome).
%For example, \citet{choi2016doctor} report that RNNs performed much worse on the less common diseases (\textit{e.g.} Klinefelter syndrome\jimeng{let's pick a different example which is not with genetic basis as it is hard to believe we can predict anything related to Klinefelter as it is genetic and patient born with it.} ) compared to frequent conditions (\textit{e.g.} hypertension) in the dataset.

The data requirement of deep learning models comes from having to assess exponential number of combinations of input features. This can be alleviated by exploiting medical ontologies that encodes hierarchical clinical constructs and relationships among medical concepts.
Fortunately, there are many well-organized ontologies in healthcare such as the International Classification of Diseases (ICD), Clinical Classifications Software (CCS) \citep{stearns2001snomed} or Systematized Nomenclature of Medicine-Clinical Terms (SNOMED-CT) \citep{healthcare2010clinical}. 
%\ecedit{Special attention has been paid to accurately describe rare conditions in such medical ontologies \citep{rath2012representation,groza2015human}. What kind of attentions? Can we be more specific?} 
%\jimeng{in future, never cite things you don't know, these \citep{rath2012representation,groza2015human} are not relevant for this paper}
Nodes (\textit{i.e.} medical concepts) close to one another in medical ontologies are likely to be associated with similar patients, allowing us to transfer knowledge among them. Therefore, proper use of medical ontologies will be helpful when we lack enough data for the nodes in the ontology to train deep learning models.
%Since the medical ontologies already contain information that is close to the ground truth, it can help us cope with the difficulty of procuring sufficient data to train deep learning models. 

\begin{figure*}[t]
\caption{The illustration of \mname. Leaf nodes (solid circles) represents a medical concept in the EHR, while the non-leaf nodes (dotted circles) represent more general concepts. 
%Note that the indices $a, b, \ldots, k$ of the codes in the DAG do not represent a specific order. 
The final representation $\gb_i$ of the leaf concept $c_i$ is computed by combining the basic embeddings $\eb_i$ of $c_i$ and $\eb_g, \eb_c$ and $\eb_a$ of   its ancestors $c_g, c_c$ and $c_a$ via an attention mechanism. The final representations form the embedding matrix $\Gb$ for all leaf concepts. 
After that, we use $\Gb$ to embed patient visit vector $\xb_t$ to a visit representation $\vb_t$, which is then fed to a neural network model to make the final prediction $\hat{\yb}_t$.}
\centering
\label{fig:graph}
\includegraphics[width=.7\textwidth]{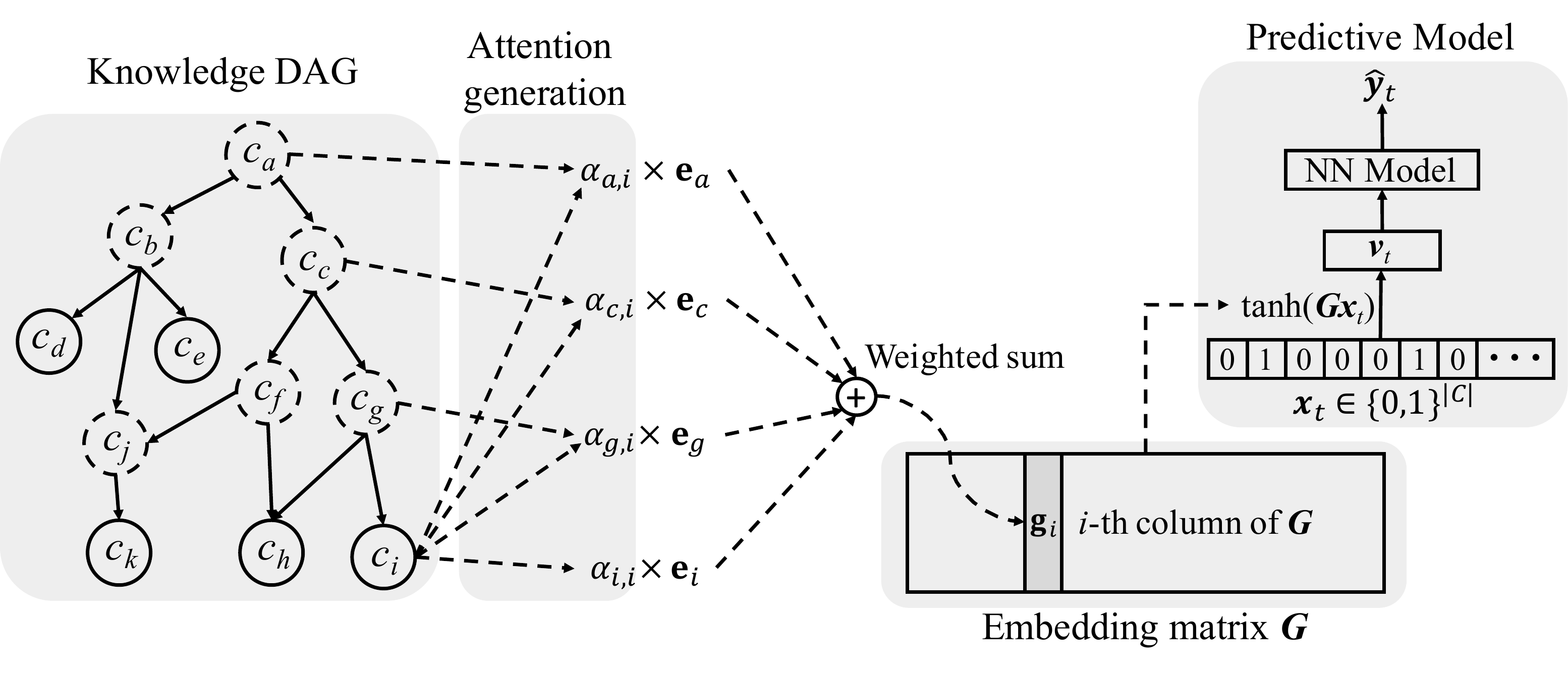}
\end{figure*}

In this work, we propose \mname, a method that infuses information from medical ontologies into deep learning models via neural attention.
Considering the frequency of a medical concept in the EHR data and its ancestors in the ontology, \mname decides the representation of the medical concept by adaptively combining its ancestors via attention mechanism.
This will not only support deep learning models to learn robust representations without large amount of data, but also learn interpretable representations that align well with the knowledge from the ontology.
%It is well-known \citep{bengio2013representation} that the predictive performance of neural network models depends on the quality of the input representation. In order to learn better input representations under the data insufficiency problem\footnote{The rare diseases and lack of training data are essentially causing the same data problem. Therefore we use the expression \textit{data insufficiency} to cover both cases.}, we propose \mname, a method to infuse such medical ontologies into neural networks via neural attention on graphs. Specifically, \mname uses medical ontology graphs to better learn the representations of medical concepts and increase the prediction performance when dealing with insufficient data or rare diseases. This is achieved by employing an attention mechanism on the ontology graph to represent a medical concept as a weighted combination of itself and its ancestors, see Figure \ref{fig:graph} for an example. 
The attention mechanism is trained in an end-to-end fashion with the neural network model that predicts the onset of disease(s). We also propose an effective initialization technique in addition to the ontological knowledge to better guide the representation learning process.

%Our approach is model-free and does not need any similarity metric on the graph to construct a graph-Laplacian regularizaer, a popular approach for using the information provided in the form of a graph \citep{}. \jimeng{we should only include this last sentence if we have experiments to support }
%Moreover, it can be used together with the graph-Laplacian regularizer as well. 
%Our method can be viewed as an adaptive medical code grouper, which rolls up the codes according to the structure of the ontology graph and the code frequency in the dataset. This facilitates learning more robust representations, thus enhancing the generalization of the algorithm under the data insufficiency problem. 

We compare predictive performance (i.e. accuracy, data needs, interpretability) of \mname to various models including the recurrent neural network (RNN) in  two sequential diagnoses prediction tasks and one heart failure (HF) prediction task.
We demonstrate that \mname is up to 10\% more accurate than the basic RNN for predicting diseases less observed in the training data.
After discussing \mname's scalability, we visualize the representations learned from various models, where \mname provides more intuitive representations by grouping similar medical concepts close to one another.
Finally, we show \mname's attention mechanism can be interpreted to understand how it assigns the right amount of attention to the ancestors of each medical concept by considering the data availability and the ontology structure.

%\vspace*{-2mm}
\section{Methodology}
\label{sec:method}
We first define the notations describing EHR data and medical ontologies, followed by a description of \mname (Section \ref{sec:attention}), the end-to-end training of the attention generation and predictive modeling (Section \ref{sec:e2e}), and the efficient initialization scheme (Section \ref{sec:emb}). %\ecedit{The last sentence can be confusing to the read. Should it be more specific? Such as "we talk about initialization technique"}

%\begin{figure}[t]
%\floatbox[{\capbeside\thisfloatsetup{capbesideposition={left,center},capbesidewidth=5cm}}]{figure}[8cm]
%{\caption{A subtree of the CCS grouper for ICD9 disease codes. $\eb_i$ represents the basic embedding vector assigned to each node and the gray text is the diagnosis that represents each node. The leaf nodes are the codes that actually appear in the EHR. (Note the ICD9 codes in the parentheses) If we assume this is the entire CCS tree, then the code vocabulary size (\textit{i.e.} the number of leaf nodes) $|\mathcal{C}|=4$ and there are 7 additional ancestor codes.}  \label{fig:graph}}
%{\quad\includegraphics[scale=0.3]{./Figs/ccs_tree4}}   
%\end{figure}

%\vspace*{-2mm}
\subsection{Basic Notation}
We denote the set of entire medical codes from the EHR as $c_1, c_2, \ldots,$  $c_{|\mathcal{C}|} \in \mathcal{C}$ with the vocabulary size $|\mathcal{C}|$. The clinical record of each patient can be viewed as a sequence of visits $V_1, \ldots, V_T$ where each visit contains a subset of medical codes $V_t \subseteq \mathcal{C}$. $V_t$ can be represented as a binary vector $\xb_t \in \{0,1\}^{|\mathcal{C}|}$ where the $i$-th element is 1 only if $V_t$ contains the code $c_i$. To avoid clutter, all algorithms will be presented for a single patient.

We assume that a given medical ontology $\mathcal{G}$ typically expresses the hierarchy of various medical concepts in the form of a \textit{parent-child} relationship, where the medical codes $\mathcal{C}$ form the leaf nodes. Ontology $\mathcal{G}$ is represented as a directed acyclic graph (DAG) whose nodes form a set $\mathcal{D} = \mathcal{C} + \mathcal{C'}$. The set $\mathcal{C'}=\{c_{|\mathcal{C}|+1}, c_{|\mathcal{C}|+2}, \ldots, c_{|\mathcal{C}|+|\mathcal{C'}|}\}$ consists of all non-leaf nodes (\textit{i.e.} ancestors of the leaf nodes), where $|\mathcal{C'}|$ represents the number of all non-leaf nodes. We use \textit{knowledge DAG} to refer to $\mathcal{G}$. A parent in the knowledge DAG $\mathcal{G}$ represents a related but more general concept over its children. Therefore, $\mathcal{G}$ provides a multi-resolution view of medical concepts with different degrees of specificity. While some ontologies are exclusively expressed as parent-child hierarchies (e.g. ICD-9, CCS), others are not. For example, in some instances SNOMED-CT also links medical concepts to causal or treatment relationships, but the majority relationships in SNOMED-CT are still parent-child.  Therefore, we focus on the parent-child relationships in this work.

%\vspace*{-2mm}
\subsection{Knowledge DAG and the Attention Mechanism}
\label{sec:attention}
\mname leverages the \textit{parent-child} relationship of $\mathcal{G}$ to learn robust representations when data volume is constrained.
\mname balances the use of ontology information in relation to data volume in determining the level of specificity for a medical concept.
When a medical concept is less observed in the data, more weight is given to its ancestors as they can be learned more accurately and offer general (coarse-grained) information about their children. 
The process of resorting to the parent concepts can be automated via the attention mechanism and the end-to-end training as described in Figure \ref{fig:graph}.
%In what follows, we describe the details of such statistical complexity control mechanism with the attention on trees.
% Therefore we pay attention to the parent-child relationship and view $\mathcal{G}$ as a tree.

In the knowledge DAG, each node $c_i$ is assigned a basic embedding vector $\eb_i \in \mathbb{R}^{m}$,
%, typically the leaf nodes are the  medical codes that appear in the actual EHR data. For example, the leaf nodes of Figure \ref{fig:graph} are the five digit ICD9 disease codes that are often used in EHR data. Note that the basic embedding vector $\eb_i$'s are assigned to leaf nodes first and the root node last. 
where $m$ represents the dimensionality. Then $\eb_1, \ldots, \eb_{|\mathcal{C}|}$ are the basic embeddings of the codes $c_1, \ldots, c_{|\mathcal{C}|}$ while $\eb_{|\mathcal{C}|+1}, \ldots, \eb_{|\mathcal{C}|+|\mathcal{C'}|}$ represent the basic embeddings of the internal nodes $c_{|\mathcal{C}|+1}, \ldots, c_{|\mathcal{C}|+|\mathcal{C'}|}$. 
The initialization of these basic embeddings is described in Section~\ref{sec:emb}. 
%As mentioned above, we aim to exploit the parent-child relationship, but we take this motivation one step further: instead of using only the parents of leaf nodes, we use a weighted average of the entire ancestors to complement what we cannot learn reliably from the given dataset. 
We formulate a leaf node's final representation as a convex combination of the basic embeddings of itself and its ancestors: 
\begin{equation}
\gb_i = \sum_{j \in \mathcal{A}(i)} \alpha_{ij} \mathbf{e}_j, \qquad \sum_{j \in \mathcal{A}(i)} \alpha_{ij} = 1,~~ \alpha_{ij} \geq 0 ~\text{ for } j\in \mathcal{A}(i), \label{eq:final_representation}
\end{equation}
where $\gb_i \in \mathbb{R}^{m}$ denotes the final representation of the code $c_i$, $\mathcal{A}(i)$ the indices of the code $c_i$ and $c_i$'s ancestors, $\eb_j$ the basic embedding of the code $c_j$ and $\alpha_{ij} \in \mathbb{R}$ the attention weight on the embedding $\eb_j$ when calculating $\gb_i$. %\citep{bahdanau2014neural} 
%The embedding dimension $m$ is selected to be the same for both the basic embedding and the final representation.
The attention weight $\alpha_{ij}$ in Eq. \eqref{eq:final_representation} is calculated by the following Softmax function,
\begin{equation}
\alpha_{ij} = \frac{\exp(f(\eb_i, \eb_j))}{\sum_{k\in \mathcal{A}(i)} \exp(f(\eb_i, \eb_k))}
\label{eq:softmax}
\end{equation}
$f(\eb_i, \eb_j)$ is a scalar value representing the compatibility between the basic embeddings of $\eb_i$ and $\eb_k$. We compute $f(\eb_i, \eb_j)$ via the following feed-forward network with a single hidden layer (MLP),
\begin{equation}
f(\eb_i, \eb_j)  = \ub_{a}^{\top}\tanh(\Wb_{a} \left[ \begin{array}{c}\eb_i \\ \eb_j\end{array}\right] + \bb_a)
\label{eq:attention_function}
\end{equation}
where $\Wb_a \in \mathbb{R}^{l \times 2m}$ is the weight matrix for the concatenation of $\eb_i$ and $\eb_j$, $\bb \in \mathbb{R}^{l}$ the bias vector, and $\ub_a \in \mathbb{R}^{l}$ the weight vector for generating the scalar value. The constant $l$ represents the dimension size of the hidden layer of $f(\cdot,\cdot)$. We always concatenate $\eb_i$ and $\eb_j$ in the child-ancestor order. Note that the compatibility function $f$ is an MLP, because MLP is well known to be a sufficient approximator for an arbitrary function, and we empirically found that our formulation performed better in our use cases than alternatives such as inner product and Bahdanau et al.'s \citep{bahdanau2014neural}. %is similar to Bahdanau et al.'s \citep{bahdanau2014neural} approach. However an MLP is well known to be a  sufficient approximator for an arbitrary function, and we empirically found that our formulation performed better in our use cases.}

\noindent{\bf Remarks:} 
The example in Figure \ref{fig:graph} is derived based on a single path from $c_i$ to $c_a$. However, the same mechanism can be applicable to multiple paths as well. For example, code $c_k$ has two paths to the root $c_a$, containing five ancestors in total.
Another scenario is where the EHR data contain both leaf codes and some ancestor codes.  We can move those ancestors present in EHR data from the set $\mathcal{C'}$ to $\mathcal{C}$ and apply the same process as Eq. \eqref{eq:final_representation} to obtain the final representations for them.

%In order to obtain the attention weights, we use the following feedforward network with a single hidden layer (MLP) to first calculate an intermediate scalar value $\beta_{ij}$.
%\begin{equation}
%f(\eb_i, \eb_j) = \beta_{ij} = \ub_{a}^{\top}\tanh(\Wb_{a} \left[ \begin{array}{c}\eb_i \\ \eb_j\end{array}\right] + \bb_a)
%\label{eq:attention_function}
%\end{equation}
%where $\Wb_a \in \mathbb{R}^{l \times 2m}$ is the weight matrix for the concatenation of $\eb_i$ and $\eb_j$, $\bb \in \mathbb{R}^{l}$ the bias vector, and $\ub_a \in \mathbb{R}^{l}$ the weight vector for generating the scalar value. The constant $l$ represents the dimension size of the hidden layer of $f(\cdot,\cdot)$\footnote{Note that we always concatenate $\eb_i$ and $\eb_j$ in the child-ancestor order.}. Once we have the scalar values $\beta_{ij}$ for all $j \in \mathcal{A}(i)$, we use the Softmax function to obtain the attention weights $\alpha_{ij}$ as follows, 
%\begin{equation}
%\alpha_{ij} = \frac{\exp(\beta_{ij})}{\sum_{k\in %\mathcal{A}(i)} \exp(\beta_{ik})}
%\label{eq:softmax}
%\end{equation}
%Then the final representation $\gb_i$ for the code $c_i$ can be calculated by Eq. (\eqref{eq:final_representation}).

%\vspace*{-2mm}
\subsection{End-to-End Training with a Predictive Model}
\label{sec:e2e}
We train the attention mechanism together with a predictive model such that the attention mechanism improves the predictive performance.
%\ecedit{Note that our method, applying attention on knowledge DAG $\mathcal{G}$, can be used for other tasks such as unsupervised representation learning. However, we select predictive modeling to facilitate rigorous evaluation of the performance gain. Are we keeping this sentence? If we are only focusing on supervised learning, we should remove this.}
By concatenating final representation $\gb_1, \gb_2, \ldots, \gb_{|\mathcal{C}|}$ of all medical codes, we have the embedding matrix %(\textit{i.e.} lookup table) 
$\Gb \in \mathcal{R}^{m \times |\mathcal{C}|}$ where $\gb_i$ is its $i$-th column of $\Gb$. 
We can then convert visit $V_t$ to a visit representation $\vb_t$ by multiplying embedding matrix $\Gb$ with multi-hot vector $\mathbf{x}_t$ indicating the clinical events in visit $V_t$ as shown in the right side of Figure \ref{fig:graph}. Finally the visit representation $\vb_t$ will be used as an input to pass to a predictive model for predicting the target label $\mathbf{y}_t$ using a neural network (NN) model. In this work, we use RNN as the choice of the NN model as the task is to perform sequential diagnoses prediction  \citep{choi2016doctor, choi2016retain} with the objective of predicting the disease codes of the next visit $V_{t+1}$ given the visit records up to the current timestep $V_1, V_2, \ldots, V_{t}$, which can be expressed as follows, 
\begin{align}
\widehat{\yb}_{t} = \widehat{\xb}_{t+1} & = \mathrm{Softmax}(\Wb \hb_{t} + \bb), \text{\quad where} \nonumber \\
\hb_1, \hb_2, \ldots, \hb_{t} & = \textrm{RNN}(\vb_1, \vb_2, \ldots, \vb_{t}; \theta_{r}), \text{\quad where}\label{eq:rnn_prediction} \\
\vb_1, \vb_2, \ldots, \vb_{t} & = \tanh(\Gb [\xb_1, \xb_2, \ldots, \xb_{t}]) \nonumber
\end{align}
% \jimeng{it seems it should be $G^T$ instead of $G$ in both the equations and in figure 1; also we should again state what $v_i$ and $x_i$ are and their size } 
where $\xb_t \in \mathbb{R}^{|\mathcal{C}|}$ denotes the $t$-th visit; $\vb_t \in \mathbb{R}^m$ the $t$-th visit representation; $\hb_t \in \mathbb{R}^{r}$ the RNN's hidden layer at $t$-th time step (\textit{i.e.} $t$-th visit); $\theta_r$ RNN's parameters; $\Wb \in \mathbb{R}^{|\mathcal{C}| \times r}$ and $\bb \in \mathbb{R}^{|\mathcal{C}|}$ the weight matrices and the bias vector of the Softmax function; $r$ denotes the dimension size of the hidden layer. We use ``$\mathrm{RNN}$'' to denote any recurrent neural network variants that can cope with the vanishing gradient problem \citep{bengio1994learning}, such as LSTM \citep{hochreiter1997long}, GRU \citep{cho2014learning}, and IRNN \citep{le2015simple}, with any varying numbers of hidden layers. The prediction loss for all time steps is calculated using the binary cross entropy as follows,
\small
\begin{equation} 
\mathcal{L}(\xb_1, \xb_2 \ldots, \xb_T) = -\frac{1}{T-1} \sum_{t=1}^{T-1} \Big( {\yb_{t}}^{\top} \log(\widehat{\yb}_{t}) + (\mathbf{1} - \yb_{t})^{\top} \log(\mathbf{1} - \widehat{\yb}_{t}) \Big) \label{eq:dpm_loss}
\end{equation}
\normalsize
 where we sum the cross entropy errors from all timestamps of $\widehat{\yb}_{t}$, $T$ denotes the number of timestamps of the visit sequence. Note that the above loss is defined for a single patient. In actual implementation, we will take the average of the individual loss for multiple patients.
 Algorithm~\ref{alg:training} describes the overall training procedure of \mname, under the assumption that we are performing the sequential diagnoses prediction task using an RNN. Note that Algorithm~\ref{alg:training} describes stochastic gradient update to avoid clutter, but it can be easily extended to other gradient based optimization such as mini-batch gradient update.
\begin{algorithm}[tb]
   \caption{\mname Optimization}
   \label{alg:training}
\begin{algorithmic}
   \STATE Randomly initialize basic embedding matrix $\Eb$, attention parameters $\ub_a, \Wb_a, \bb_a$, RNN parameter $\theta_r$, softmax parameters $\Wb, \bb$.
   \REPEAT
   \STATE Update $\Eb$ with GloVe objective function (see Section \ref{sec:emb})
   \UNTIL{convergence}
   \REPEAT
   \STATE $\Xb \leftarrow$ random patient from dataset
   \FOR{visit $V_t$ {\bfseries in} $\Xb$}
   \FOR{code $c_i$ {\bfseries in} $V_t$}
   \STATE Refer $\mathcal{G}$ to find $c_i$'s ancestors $C'$
   \FOR{code $c_j$ {\bfseries in} $C'$}
   \STATE Calculate attention weight $\alpha_{ij}$ using Eq. \eqref{eq:softmax}.
   \ENDFOR
   \STATE Obtain final representation $\gb_i$ using Eq. \eqref{eq:final_representation}.
   \ENDFOR
   \STATE $\vb_t \leftarrow \mbox{tanh}(\sum_{i:c_i \in V_t} \gb_i)$ %Obtain $\vb_t$ by summing all $\gb_i$ and applying tanh. \jsedit{Let's just use the math equation here to be more concise}
   \STATE Make prediction $\widehat{\yb}_{t}$ using Eq. \eqref{eq:rnn_prediction}
   \ENDFOR
   \STATE Calculate prediction loss $\mathcal{L}$ using Eq .\eqref{eq:dpm_loss}
   \STATE Update parameters according to the gradient of $\mathcal{L}$
   \UNTIL{convergence}
\end{algorithmic}
\end{algorithm}

%Overall, \mname forces use of ancestor knowledge embedded in $\mathcal{G}$ so that RNN maximizes predictive performance as it learns the basic embeddings $\eb_i$ and the attention mechanism to control medical concept specificity and produce clinically interpretable representations.
%Note that, however, \mname can be used with any neural network model that can be trained via backpropagation. \ecedit{In the experiments, we use both RNN and MLP to demonstrate this.}

%\vspace*{-2mm}
\subsection{Initializing Basic Embeddings}
\label{sec:emb}
The attention generation mechanism in Section \ref{sec:attention} requires basic embeddings $\eb_i$ of each node in the knowledge DAG. The basic embeddings of ancestors, however, pose a difficulty because they are often not observed in the data. %, and require use of a more principled approach than random initialization. 
To properly initialize them, we use co-occurrence information to learn the basic embeddings of medical codes and their ancestors. Co-occurrence has proven to be an important source of information when learning representations of words or medical concepts \citep{mikolov2013distributed, choi2016multi, choi2016learning}. To train the basic embeddings, we employ GloVe \citep{pennington2014glove}, which uses the global co-occurrence matrix of words to learn their representations. %As shown in Figure \ref{fig:co-occur}, 
In our case, the co-occurrence matrix of the codes and the ancestors was generated by counting the co-occurrences within each visit $V_t$, where we augment each visit with the ancestors of the codes in the visit.
\begin{figure}
%\floatbox[{\capbeside\thisfloatsetup{capbesideposition={left,center},capbesidewidth=7cm}}]{figure}[\FBwidth]	
\centering
\caption{Creating the co-occurrence matrix together with the ancestors. The $n$-th ancestors are the group of nodes that are $n$ hops away from any leaf node in $\mathcal{G}$. Here we exclude the root node, which will be just a single row (column).}
\label{fig:co-occur}
\includegraphics[scale=0.28]{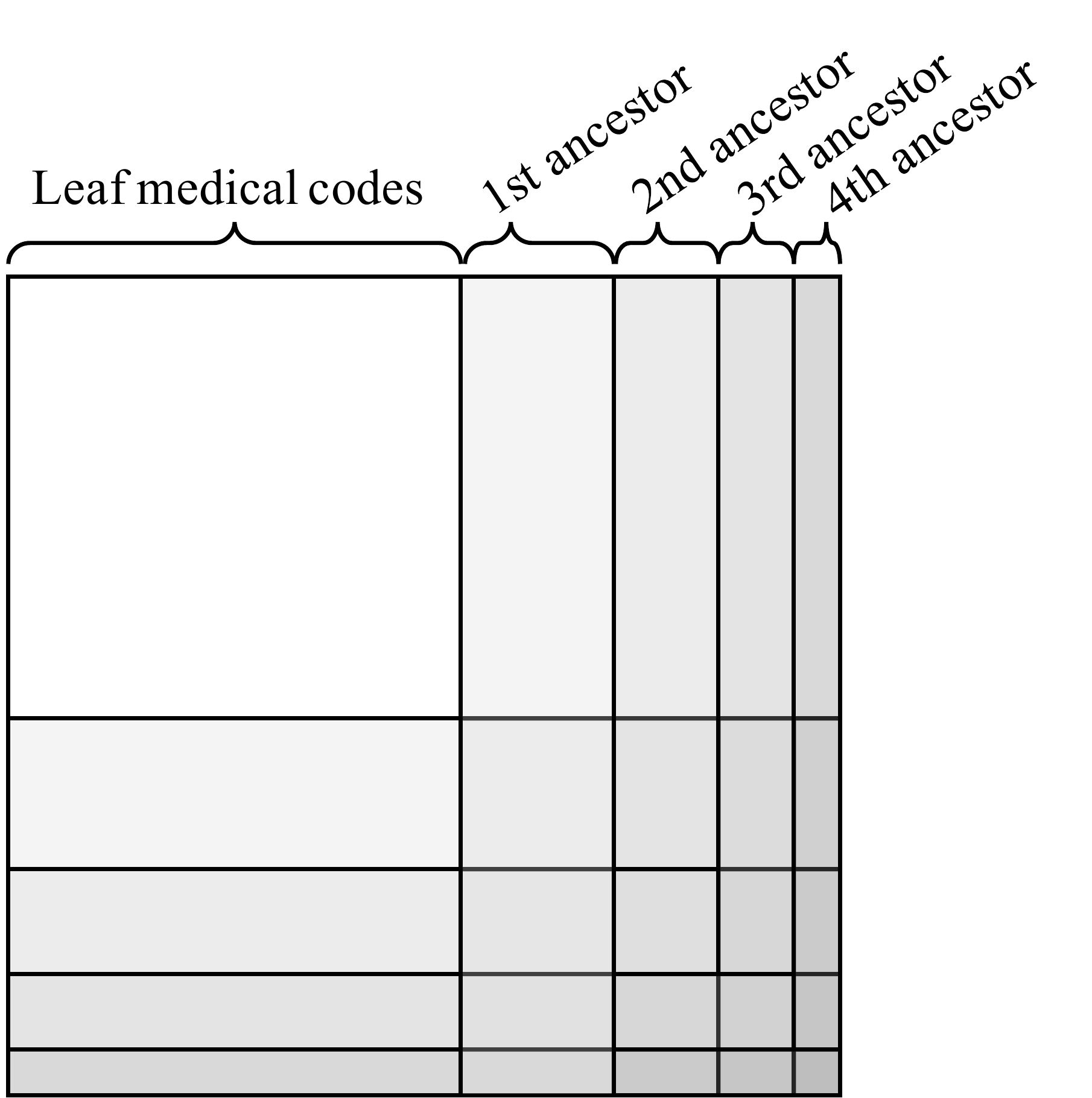}
\end{figure}
We describe the details of the initialization algorithm with an example. We borrow the parent-child relationships from the knowledge DAG of Figure \ref{fig:graph}. Given a visit $V_t$,

\begin{equation*}
V_t = \{c_d, c_i, c_k\}
\end{equation*}
we augment it with the ancestors of all the codes to obtain the augmented visit $V'_t$,
\begin{equation*}
V'_t = \{c_d, \underline{c_b}, \underline{c_a}, c_i, \underline{c_g}, \underline{c_c}, \underline{c_a}, c_k, \underline{c_j}, \underline{c_f}, \underline{c_c}, \underline{c_b}, \underline{c_a}\}
\end{equation*}
where the augmented ancestors are underlined. Note that a single ancestor can appear multiple times in $V'_t$. In fact, the higher the ancestor is in the knowledge DAG, the more times it is likely to appear in $V'_t$. We count the co-occurrence of two codes in $V'_t$ as follows,
\begin{equation*}
co\mbox{-}occurrence(c_i, c_j, V'_t) = count(c_i, V'_t) \times count(c_j, V'_t)
\end{equation*}
where $count(c_i, V'_t)$ is the number of times the code $c_i$ appears in the augmented visit $V'_t$. For example, the co-occurrence between the leaf code $c_i$ and the root $c_a$ is 3. However, the co-occurrence between the ancestor $c_c$ and the root $c_a$ is 6. Therefore our algorithm will make the higher ancestor codes, which are more general concepts, have more involvement in all medical events (\textit{i.e.} visits), which is natural in healthcare application as those general concepts are often reliable. We repeat this calculation for all pairs of codes in all augmented visits of all patients to obtain the co-occurrence matrix $\Mb \in \mathbb{R}^{|\mathcal{D}| \times |\mathcal{D}|}$ depicted by Figure \ref{fig:co-occur}. For training the embedding vectors $\eb_i$'s using $\Mb$, we minimize the following loss function as described in \citet{pennington2014glove}.
\begin{align*}
J & = \sum_{i,j=1}^{|\mathcal{D}|} f(\Mb_{ij}) (\eb_i^{\top} \eb_j + b_i + b_j - \log \Mb_{ij})^2\\
\mbox{where } & f(x) = \begin{cases} 
(x < x_{max})^{\alpha} & \text{if $x < x_{max}$} \\
1 & \text{otherwise}
\end{cases}
\end{align*}
where the hyperparameters $x_{max}$ and $\alpha$ are respectively set to $100$ and $0.75$ as the original paper \citep{pennington2014glove}.
Note that, after the initialization, the basic embeddings $\eb_i$'s of both leaf nodes (\textit{i.e.} medical codes) and non-leaf nodes (\textit{i.e.} ancestors) are fine-tuned during model training via backpropagation.%, since the error signal flows from the output $\widehat{\yb}_{t}$ to the final representations $\gb_i$'s which are convex combinations of $\eb_i$'s.

%\vspace*{-2mm}
\section{Experiments}
\label{sec:exp}
We conduct three experiments to determine if \mname offered superior prediction performance when facing data insufficiency. We first describe the experimental setup followed by results comparing predictive performance of \mname with various baseline models. After discussing \mname's scalability, we qualitatively evaluate the interpretability of the resulting representation.  
The source code of \mname is publicly available at \url{https://github.com/mp2893/gram}.

%\vspace*{-2mm}
\subsection{Experiment Setup}
\begin{table}[t]
\caption{Basic statistics of Sutter PAMF, MIMIC-III and Sutter heart failure (HF) cohort.}
\label{tab:data_stats}
% \small
\vspace*{-2mm}
\centering
\begin{footnotesize}
\begin{tabular}{@{}l|c|c|c@{}}
\textbf{Dataset} & \textbf{Sutter PAMF} & \textbf{MIMIC-III} & \textbf{Sutter HF cohort} \\
\toprule
\# of patients & 258,555$^{\dagger}$ &7,499$^{\dagger}$ &30,727$^{\dagger}$ (3,408 cases)\\ 
\# of visits & 13,920,759 & 19,911 & 572,551\\
Avg. \# of visits per patient & 53.8 & 2.66 & 38.38\\
\# of unique ICD9 codes & 10,437 & 4,893 & 5,689\\
Avg. \# of codes per visit & 1.98 & 13.1 & 2.06\\
Max \# of codes per visit & 54 & 39 & 29\\
\bottomrule
\multicolumn{4}{c}{$\dagger$ \small For all datasets, we chose patients who made at least two visits.}
\end{tabular}
\end{footnotesize}
\vspace{-3mm}
\end{table}
\textbf{Prediction tasks and source of data:} We conduct the sequential diagnoses prediction (SDP) tasks on two datasets, which aim at predicting all diagnosis categories in the next visit, and a heart failure (HF) prediction task on one dataset, which is a binary prediction task for predicting a future HF onset where the prediction is made only once at the last visit $\xb_T$. %The key difference between these two tasks is that prediction target for the former can already occur in patient's prior visits while the prediction target for the latter is a new diagnosis of HF that has not appeared before.
\newline
Two sequential diagnoses predictions (SDP) are respectively conducted using two datasets:
1) Sutter Palo Alto Medical Foundation (PAMF) dataset, which consists of 18-years longitudinal medical records of 258K patients between age 50 and 90. This will determine \mname's performance for general adult population with long visit records. 
2) MIMIC-III dataset \citep{johnson2016mimic, goldberger2000physiobank}, which is a publicly available dataset consisting of medical records of 7.5K intensive care unit (ICU) patients over 11 years. This will determine \mname's performance for high-risk patients with very short visit records. We utilize all the patients with at least 2 visits. We prepared the true labels $\yb_{t}$ by grouping the ICD9 codes into 283 groups using CCS single-level diagnosis grouper\footnote{https://www.hcup-us.ahrq.gov/toolssoftware/ccs/AppendixASingleDX.txt}. 
This is to improve the training speed and predictive performance for easier analysis, while preserving sufficient granularity for each diagnosis. Each diagnosis code's varying frequency in the training data can be viewed as different degrees of data insufficiency. We calculate \textit{Accuracy@k} for each of  CCS single-level diagnosis codes such that, given a visit $V_t$, we get 1 if the target diagnosis is in the top $k$ guesses and 0 otherwise.
\newline
We conduct HF prediction on Sutter heart failure (HF) cohort, which is a subset of Sutter PAMF data for a heart failure onset prediction study with 3.4K HF cases chosen by a set of criteria described in \citet{vijayakrishnan2014prevalence,gurwitz2013contemporary} and 27K matching controls chosen by a set of criteria described in \citet{choi2016using}. This will determine \mname's performance for a different prediction task where we predict the onset of one specific condition. We randomly downsample the training data to create different degrees of data insufficiency. We use area under the ROC curve (AUC) to measure the performance.
\newline
%For all three tasks, the features are extracted from ICD9 diagnosis codes in the EHR data. 
A summary of the datasets are provided in Table~\ref{tab:data_stats}.We used CCS multi-level diagnoses hierarchy\footnote{https://www.hcup-us.ahrq.gov/toolssoftware/ccs/AppendixCMultiDX.txt} as our knowledge DAG $\mathcal{G}$. We also tested the ICD9 code hierarchy\footnote{http://www.icd9data.com/2015/Volume1/default.htm}, but the performance was similar to using CCS multi-level hierarchy. %We left using SNOMED-CT for future work because it uses its own diagnosis codes, which have $m$-to-$n$ mapping relations with ICD9 diagnosis codes.
For all three tasks, we randomly divide the dataset into the training, validation and test set by .75:.10:.15 ratio, and use the validation set to tune the hyper-parameters. Further details regarding the hyper-parameter tuning are provided below. The test set performance is reported in the paper.
% \jimeng{why don't we use other sources such as medication and procedure? It seems we can also include those sources as well which may improve the prediction accuracy further.}
% Using additional information such as medication orders or procedure orders can increase predictive performance, but in this paper we focus on comparing how \mname and other baselines perform under data insufficiency.

\noindent \textbf{Implementation details:} We implemented \mname with Theano 0.8.2 \citep{team2016theano}. For training models, we used Adadelta \citep{zeiler2012adadelta} with a mini-batch of 100 patients, on a machine equipped with Intel Xeon E5-2640, 256GB RAM, four Nvidia Titan X's and CUDA 7.5. 

\noindent \textbf{Models for comparison}  are the following. The first two  {\texttt{GRAM}+} and {\texttt{GRAM}} are the proposed methods and the rest are baselines. Hyper-parameter tuning is configured so that the number of parameters for the baselines would be comparable to \texttt{GRAM}'s. Further details are provided below.
\vspace*{-1mm}
\begin{itemize}[leftmargin=5.5mm]
\item \textbf{\mname:} Input sequence $\xb_1, \ldots, \xb_T$ is first transformed by the embedding matrix $\Gb$, then fed to the GRU with a single hidden layer, which in turn makes the prediction, as described by Eq. \eqref{eq:rnn_prediction}. The basic embeddings $\eb_i$'s are randomly initialized.
\item \textbf{\texttt{GRAM}+:} We use the same setup as \textbf{\texttt{GRAM}}, but the basic embeddings $\eb_i$'s are initialized according to Section \ref{sec:emb}.
%\vspace*{-1mm}
\item
\textbf{RandomDAG:} We use the same setup as \textbf{\mname}, but each leaf concept has five randomly assigned ancestors from the CCS multi-level hierarchy to test the effect of correct domain knowledge.% as opposed to simply increasing the number of model parameters.
%\vspace*{-1mm}
\item
\textbf{RNN}: Input $\xb_t$ is transformed by an embedding matrix $\Wb_{emb} \in \mathbb{R}^{k \times |\mathcal{C}|}$, then fed to the GRU with a single hidden layer. The embedding size $k$ is a hyper-parameter. $\Wb_{emb}$ is randomly initialized and trained together with the GRU.
%\vspace*{-1mm}
\item
\textbf{RNN+:} We use the \textbf{RNN} model with the same setup as before, but we initialize the embedding matrix $\Wb_{emb}$ with GloVe vectors trained only with the co-occurrence of leaf concepts. This is to compare \mname with a similar weight initialization technique.
%\vspace*{-1mm}
\item
\textbf{SimpleRollUp:} We use the \textbf{RNN} model with the same setup as before. But for input $\xb_t$, we replace all diagnosis codes with their direct parent codes in the CCS multi-level hierarchy, giving us 578, 526 and 517 input codes respectively for Sutter data, MIMIC-III and Sutter HF cohort. This is to compare the performance of \mname with a common grouping technique.  
%\vspace*{-1mm}
\item
\textbf{RollUpRare:} We use the \textbf{RNN} model with the same setup as before, but we replace any diagnosis code whose frequency is less than a certain threshold in the dataset with its direct parent. We set the threshold to 100 for Sutter data and Sutter HF cohort, and 10 for MIMIC-III, giving us 4,408, 935 and 1,538 input codes respectively for Sutter data, MIMIC-III and Sutter HF cohort. This is an intuitive way of dealing with infrequent medical codes.
\end{itemize}
%\jimeng{remove the line space between items in itemize}

\noindent \textbf{Hyper-parameter Tuning:}
We define five hyper-parameters for \mname:
\vspace*{-1mm}
\begin{itemize}[leftmargin=5.5mm]
\item dimensionality $m$ of the basic embedding $\eb_i$: [100, 200, 300, 400, 500]
\item dimensionality $r$ of the RNN hidden layer $\hb_t$ from Eq. \eqref{eq:rnn_prediction}: [100, 200, 300, 400, 500]
\item dimensionality $l$ of $\Wb_a$ and $\bb_a$ from Eq. \eqref{eq:attention_function}: [100, 200, 300, 400, 500]
\item $L_2$ regularization coefficient for all weights except RNN weights: [0.1, 0.01, 0.001, 0.0001]
\item dropout rate for the dropout on the RNN hidden layer: [0.0, 0.2, 0.4, 0.6, 0.8]
\end{itemize}
We performed 100 iterations of the random search by using the above ranges for each of the three prediction experiments. In order to fairly compare the model performances, we matched the number of model parameters to be similar for all baseline methods.
To facilitate reproducibility, final hyper-parameter settings we used for all models for each prediction experiments are described at the source code repository, \url{https://github.com/mp2893/gram}, along with the detailed steps we used to tune the hyper-parameters.

%For sequential diagnoses prediction on Sutter data, we used 10\% of the training data to tune the hyper-parameters to balance the time and search space.
%To match the baselines' number of parameters to \mname's, we add 550 to the list of $m$'s possible values. This will make the baseline's largest possible number of parameters comparable to the \mname's largest possible number of parameters.\jsedit{Rephrase the previous 2 sentences: Shall we just say ``We match the similar number of model parameters in all the baseline methods''?}

%For SimpleRollUp and RollUpRare, the number of input codes is smaller than other models due to the grouping. Therefore, to match their largest possible number of parameters to \mname's, we need to add much larger values to $m$. However, after preliminary experiments, as expected, setting $m$ to too large a value degraded the performance due to overfitting. Since the number of input codes decreased due to the grouping, increasing the dimensionality of $\eb_i$ is not a logical thing to do. Therefore, for SimpleRollUp and RollUpRare, we use the same list of values for $m$ as other baselines.\jsedit{you mean as GRAM? i thought other baselines have more parameters added? a bit confusing here}

%\vspace*{-2mm}
\subsection{Prediction performance and scalability}
\label{ssec:exp_prediction}
%\begin{figure}[t]
%\vspace{-0.1in}
%\centering
%\begin{subfigure}{\textwidth}
%  \centering
%  \includegraphics[width=.9\linewidth]{Figs/legend}
%\end{subfigure}
%
%\begin{subfigure}{.33\textwidth}
%  \centering\captionsetup{width=.9\linewidth}%
%  \includegraphics[width=\linewidth]{Figs/dpm_sutter}
%  \vspace*{-5mm}
%  \caption{\textit{Accuracy@5} of sequential diagnoses prediction on Sutter data}
%  \label{fig:dpm_sutter}
%\end{subfigure}%
%\begin{subfigure}{.33\textwidth}
%  \centering\captionsetup{width=.9\linewidth}%
%  \includegraphics[width=\linewidth]{Figs/dpm_mimic}
%  \vspace*{-5mm}
%  \caption{\textit{Accuracy@20} of sequential diagnoses prediction on MIMIC}
%  \label{fig:dpm_mimic}
%\end{subfigure}%
%\begin{subfigure}{.33\textwidth}
%  \centering\captionsetup{width=.9\linewidth}%
%  \includegraphics[width=\linewidth]{Figs/hf_sutter}
%  \vspace*{-5mm}
%  \caption{AUC of HF onset prediction on Sutter HF cohort}
%  \label{fig:hf_sutter}
%\end{subfigure}
%\includegraphics[width=1.0\textwidth]{./Figs/gram_fig2}
%\vspace{-0.25in}
%\caption{Performance of three prediction tasks. The x-axis of (a) and (b) represents the labels grouped by  the percentile of their frequencies in the training data in non-decreasing order. For (c), we vary the size of the training data to train the models. (b) uses \textit{Accuracy@20} because MIMIC-III has a large average number of codes per visit (see Table \ref{tab:data_stats}).}
%\label{fig:predictions}
%\end{figure}
\begin{table*}[ht]
%\vspace{-0.1in}
\centering
\setlength\tabcolsep{3pt}
%\vspace*{-2mm}
\begin{subtable}{.9\textwidth}
\centering
%\captionsetup{width=.8\linewidth}%
%\small
\begin{tabular}{@{}l|c|c|c|c|c@{}}
Model & 0-20 & 20-40 & 40-60 & 60-80 & 80-100\\ 
\toprule
GRAM+&\textbf{0.0150}&\textbf{0.3242}&0.4325&0.4238&0.4903\\
GRAM&0.0042&0.2987&0.4224&0.4193&0.4895\\
RandomDAG&0.0050&0.2700&0.4010&0.4059&0.4853\\
RNN+&0.0069&0.2742&0.4140&0.4212&\textbf{0.4959}\\
RNN&0.0080&0.2691&0.4134&0.4227&0.4951\\
SimpleRollUp&0.0085&0.3078&\textbf{0.4369}&\textbf{0.4330}&0.4924\\
RollUpRare&0.0062&0.2768&0.4176&0.4226&0.4956\\
\bottomrule
\end{tabular}
%\normalsize
\caption{\textit{Accuracy@5} of sequential diagnoses prediction on Sutter data}
\label{tab:sdp_sutter}
\vspace*{3mm}
\end{subtable}

\begin{subtable}{.9\textwidth}
\centering
%\captionsetup{width=.8\linewidth}%
%\small
\begin{tabular}{@{}l|c|c|c|c|c@{}}
Model & 0-20 & 20-40 & 40-60 & 60-80 & 80-100\\ 
\toprule
GRAM+&\textbf{0.0672}&\textbf{0.1787}&\textbf{0.2644}&0.2490&0.6267\\
GRAM&0.0556&0.1016&0.1935&0.2296&0.6363\\
RandomDAG&0.0329&0.0708&0.1346&0.1512&0.4494\\
RNN+&0.0454&0.0843&0.2080&0.2494&0.6239\\
RNN&0.0454&0.0731&0.1804&0.2371&0.6243\\
SimpleRollUp&0.0578&0.1328&0.2455&\textbf{0.2667}&\textbf{0.6387}\\
RollUpRare&0.0454&0.0653&0.1843&0.2364&0.6277\\
\bottomrule
\end{tabular}
%\normalsize
\caption{\textit{Accuracy@20} of sequential diagnoses prediction on MIMIC-III}
\label{tab:sdp_mimic}
\vspace*{3mm}
\end{subtable}

\begin{subtable}{1.\textwidth}
\centering
%\small
\begin{tabular}{l|c|c|c|c|c|c|c|c|c|c}
Model & 10\% & 20\% & 30\% & 40\% & 50\% & 60\% & 70\% & 80\% & 90\% & 100\%\\ 
\toprule
GRAM+ & 0.7970 & \textbf{0.8223} & 0.8307 & \textbf{0.8332} & \textbf{0.8389} & \textbf{0.8404} & \textbf{0.8452} & \textbf{0.8456} & \textbf{0.8447} & \textbf{0.8448} \\
GRAM & \textbf{0.7981} & 0.8217 & \textbf{0.8340} & \textbf{0.8332} & 0.8372 & 0.8377 & 0.8440 & 0.8431 & 0.8430 & 0.8447 \\
RandomDAG&0.7644&0.7882&0.7986&0.8070&0.8143&0.8185&0.8274&0.8312&0.8254&0.8226\\
RNN+&0.7930&0.8117&0.8162&0.8215&0.8261&0.8333&0.8343&0.8353&0.8345&0.8335\\
RNN&0.7811&0.7942&0.8066&0.8111&0.8156&0.8207&0.8258&0.8278&0.8297&0.8314\\
SimpleRollUp&0.7799&0.8022&0.8108&0.8133&0.8177&0.8207&0.8223&0.8272&0.8269&0.8258\\
RollUpRare&0.7830&0.8067&0.8064&0.8119&0.8211&0.8202&0.8262&0.8296&0.8307&0.8291\\
\bottomrule
\end{tabular}
%\normalsize
\caption{AUC of HF onset prediction on Sutter HF cohort}
\label{tab:hf_sutter}
\end{subtable}

\caption{Performance of three prediction tasks. The x-axis of (a) and (b) represents the labels grouped by  the percentile of their frequencies in the training data in non-decreasing order. 0-20 are the most rare diagnosese while 80-100 are the most common ones. (b) uses \textit{Accuracy@20} because MIMIC-III has a large average number of codes per visit (see Table \ref{tab:data_stats}). For (c), we vary the size of the training data to train the models.}
\label{tab:predictions}
%\vspace*{-5mm}
\end{table*}
Tables \ref{tab:sdp_sutter} and \ref{tab:sdp_mimic} show the sequential diagnoses prediction performance on Sutter data and MIMIC-III. Both figures show that \texttt{GRAM}+ outperforms other models when predicting labels with significant data insufficiency (\textit{i.e.} less observed in the training data).%, up to 40th percentile in Sutter data and up to 60th percentile in MIMIC-III. 
 The performance gain is greater for MIMIC-III, where \texttt{GRAM}+ outperforms the basic RNN by 10\% in the 20th-40th percentile range. 
This seems to come from the fact that MIMIC patients on average have significantly shorter visit history than Sutter patients, with much more codes received per visit. Such short sequences make it difficult for the RNN to learn and predict diagnoses sequence.
%This is because MIMIC-III is much smaller than Sutter data, which amplifies the data insufficiency problem. \ecedit{Moreover, MIMIC-III contains more diverse infrequent diseases occurring in ICU, as opposed to Sutter data where the data are collected from only outpatient encounters.} 
The performance difference between \texttt{GRAM}+ and \mname suggests that our proposed initialization scheme of the basic embeddings $\eb_i$ is important for sequential  diagnosis prediction. %\ecedit{It seems that sequential diagnoses prediction on MIMIC-III benefits more from the initialization scheme due to its richer co-occurrence information (average 13.1 codes per visit).} Additional results from varying the $k$ of \textit{Accuracy@k} are discussed in Appendix \ref{appendix:prediction}.

Table \ref{tab:hf_sutter} shows the HF prediction performance on Sutter HF cohort. \texttt{GRAM} and \texttt{GRAM}+ consistently outperforms other baselines (except RNN+) by 3$\sim$4\% AUC, and RNN+ by maximum 1.8\% AUC. These differences are quite significant given that the AUC is already in the mid-80s, a high value for HF prediction, cf. \citep{choi2016using}. Note that, for GRAM+ and RNN+, we used the downsampled training data to initialize the basic embeddings $\eb_i$'s and the embedding matrix $\Wb_{emb}$ with GloVe, respectively. The result shows that the initialization scheme of the basic embeddings in \texttt{GRAM}+ gives limited improvement over \texttt{GRAM}. This stems from the different natures of the two prediction tasks. While the goal of HF prediction is to predict a binary label for the entire visit sequence, the goal of sequential diagnosis prediction is to predict the co-occurring diagnosis codes at every visit. Therefore the co-occurrence information infused by the initialized embedding scheme is more beneficial to sequential diagnosis prediction. Additionally, this benefit is associated with the natures of the two prediction tasks than the datasets used for the prediction tasks. Because the initialized embedding shows different degrees of improvement as shown by Tables \ref{tab:sdp_sutter} and \ref{tab:hf_sutter}, when Sutter HF cohort is a subset of Sutter PAMF, thus having similar characteristics.
%\subsection{\ecedit{Scalability of \mname}}
%\label{ssec:exp_scalability}
\begin{table}[t]
\vspace{-0.1in}
\caption{Scalablity result in per epoch training time in second (the number of epochs needed). SDP stands for Sequential Diagnoses Prediction }%The number in the parenthesis is the number of epochs taken to train a model with the lowest validation loss.}
\vspace*{-3mm}
\label{tab:training_time}
% \small
\centering
\begin{normalsize}
\begin{tabular}{l|c|c|c}
\textbf{Model} & \begin{tabular}{c} SDP \\ (Sutter data) \end{tabular} & \begin{tabular}{c}  SDP \\ (MIMIC-III) \end{tabular} & \begin{tabular}{c} HF prediction \\ (Sutter HF cohort) \end{tabular}\\
\toprule
\texttt{GRAM} & 525s (39 epochs) & 2s (11 epochs) & 12s (7 epochs) \\ 
RNN & 352s (24 epochs) & 1s (6 epochs) & 8s (5 epochs)
%SimpleRollUp & 165s (24) & 1s (6) & 4s (5) \\
%RollUpRare & 249s (24) & 1s (6) & 4s (5)
\\
\bottomrule
\end{tabular}
\end{normalsize}
\vspace{-5.3mm}
\end{table}
Overall, \mname showed superior predictive performance under data insufficiency in three different experiments, demonstrating its general applicability in clinical predictive modeling. Now we briefly discuss the scalability of \mname by comparing its training time to RNN's. Table \ref{tab:training_time} shows the number of seconds taken for the two models to train for a single epoch for each predictive modeling task. \texttt{GRAM}+ and RNN+ showed the similar behavior as \mname and RNN. 
%We exclude the training times of SimpleRollUp and RollUpRare as their input disease codes are grouped, making a comparison with \mname meaningless. We show the training times of the models used in Section \ref{ssec:exp_prediction} where the two models' number of parameters are comparable except for disease progression modeling on MIMIC-III, in which case RNN used slightly more parameters.
\mname takes approximately 50\% more time to train for a single epoch for all prediction tasks. This stems from calculating attention weights and the final representations $\gb_i$ for all medical codes. \mname also generally takes about 50\% more epochs to reach to the model with the lowest validation loss. This is due to optimizing an extra MLP model that generates the attention weights. Overall, use of \mname adds a manageable amount of overhead in training time to the plain RNN. %, making it a practical method to cope with data insufficiency.

%\vspace*{-2mm}
\subsection{Qualitative evaluation of interpretable representations}
\label{ssec:exp_tsne}
\begin{figure*}[h]
\centering
\vspace*{-5mm}
\begin{subfigure}{\textwidth}
  \centering\captionsetup{width=.9\linewidth}%
  \includegraphics[width=.85\linewidth]{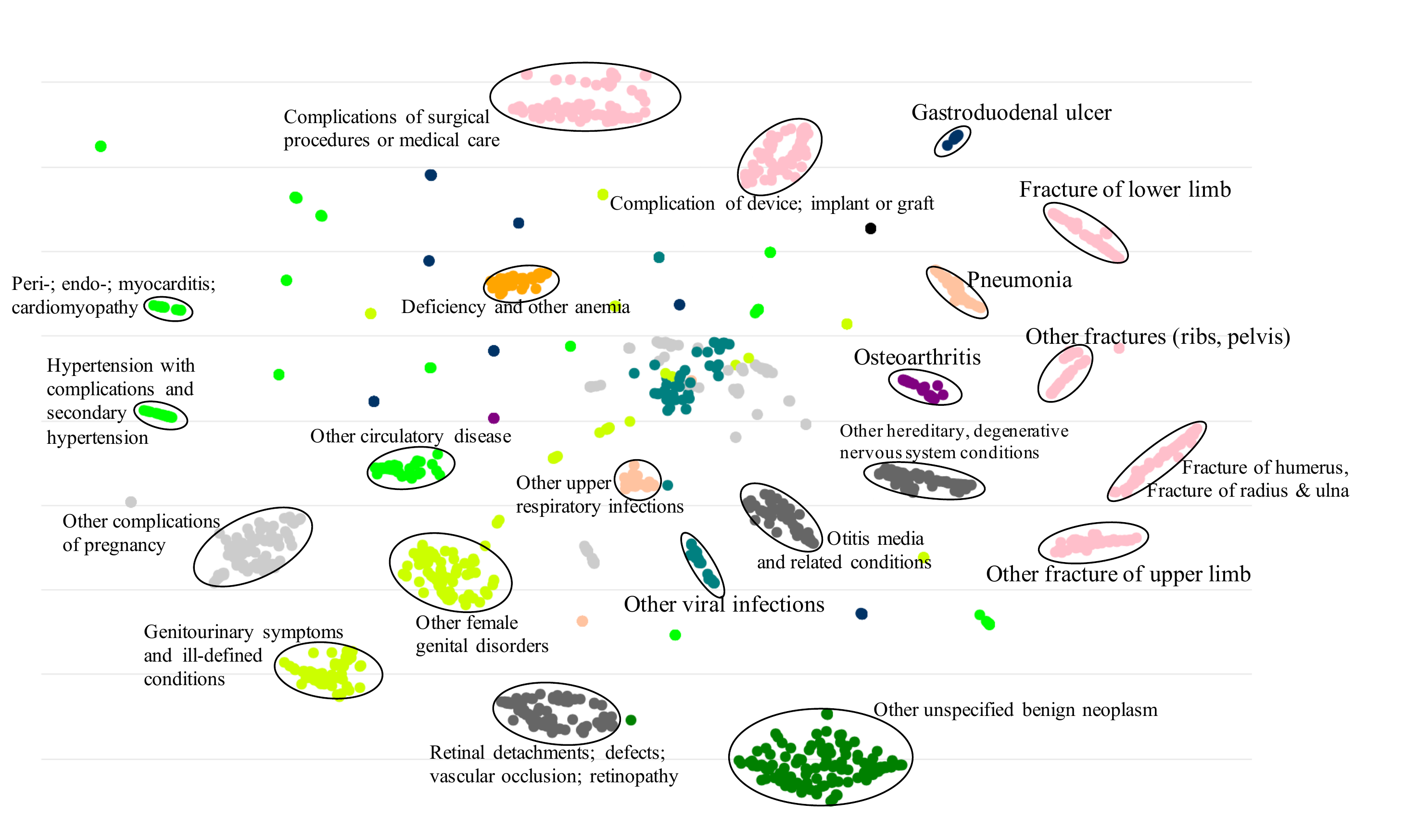}
  \vspace*{-3mm}
  \caption{Scatterplot of the final representations $\gb_i$'s of \texttt{GRAM}+}
  \label{fig:tsne_alg_glove}
\vspace{-0.01in}
\end{subfigure}

\begin{subfigure}{.5\textwidth}
  \centering\captionsetup{width=.9\linewidth}%
  \includegraphics[width=.55\linewidth]{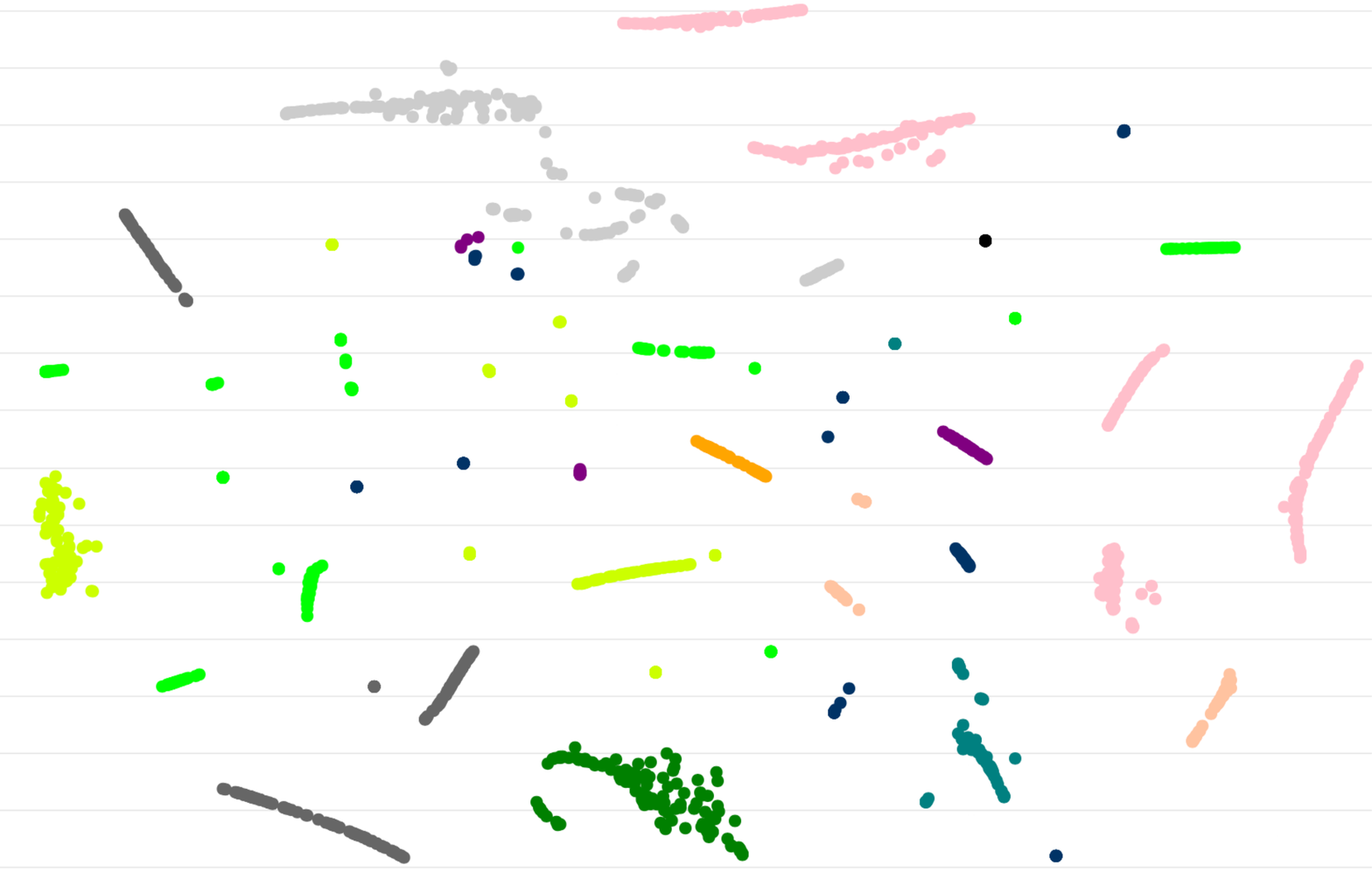}
  \vspace*{-2mm}
  \caption{Scatterplot of the final representations $\gb_i$'s of \mname}
  \label{fig:tsne_alg}
\end{subfigure}%
\begin{subfigure}{.5\textwidth}
  \centering\captionsetup{width=.9\linewidth}%
  \includegraphics[width=.55\linewidth]{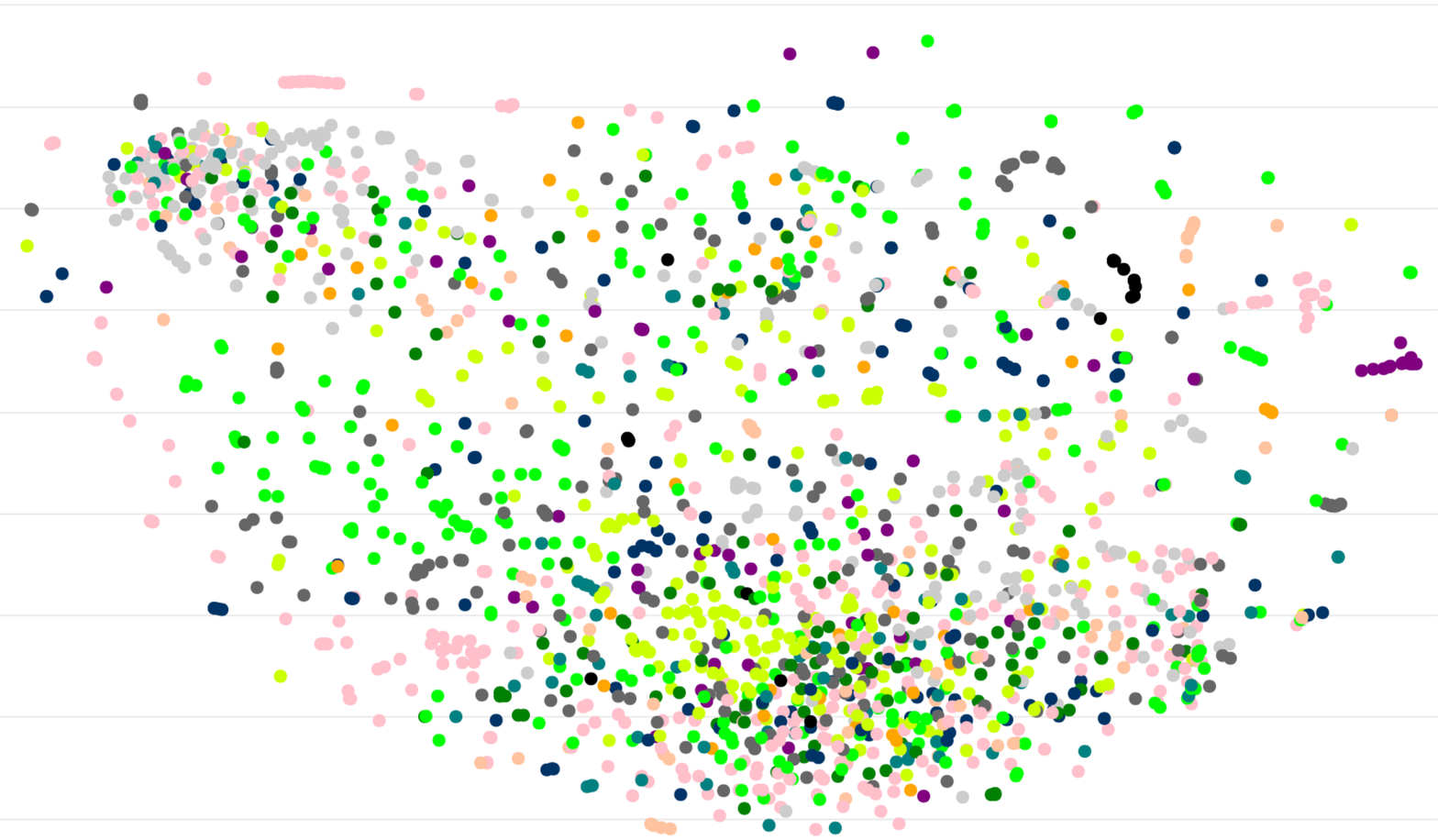}
  \vspace*{-2mm}
  \caption{Scatterplot of the trained embedding matrix $\Wb_{emb}$ of RNN+}
  \label{fig:tsne_rnn_glove}
\end{subfigure}

\begin{subfigure}{.5\textwidth}
  \centering\captionsetup{width=.9\linewidth}%
  \includegraphics[width=.55\linewidth]{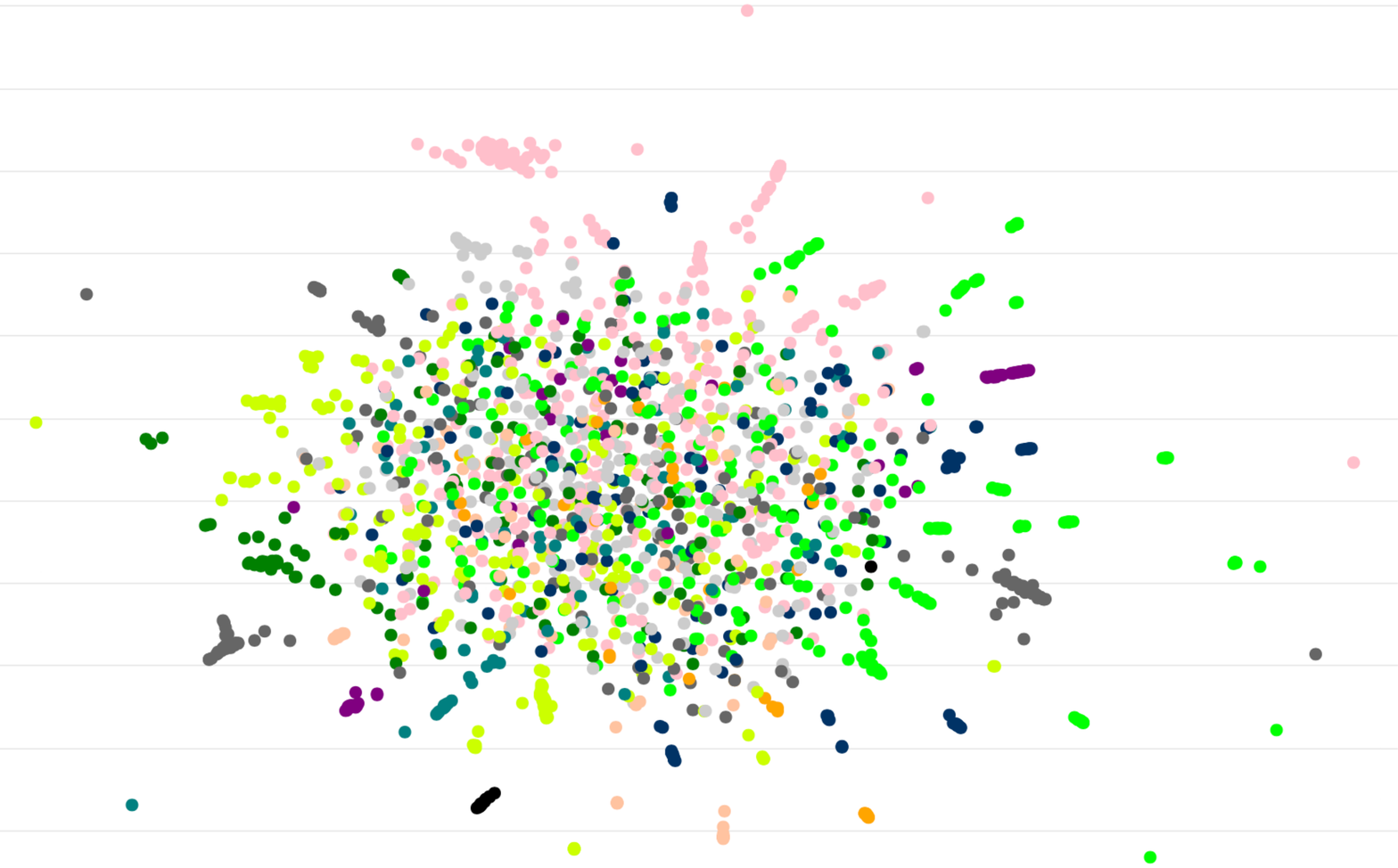}
  \vspace*{-2mm}
  \caption{Scatterplot of the trained embedding matrix $\Wb_{emb}$ of RNN}
  \label{fig:tsne_rnn}
\end{subfigure}%
\begin{subfigure}{.5\textwidth}
  \centering\captionsetup{width=.9\linewidth}%
  \includegraphics[width=.55\linewidth]{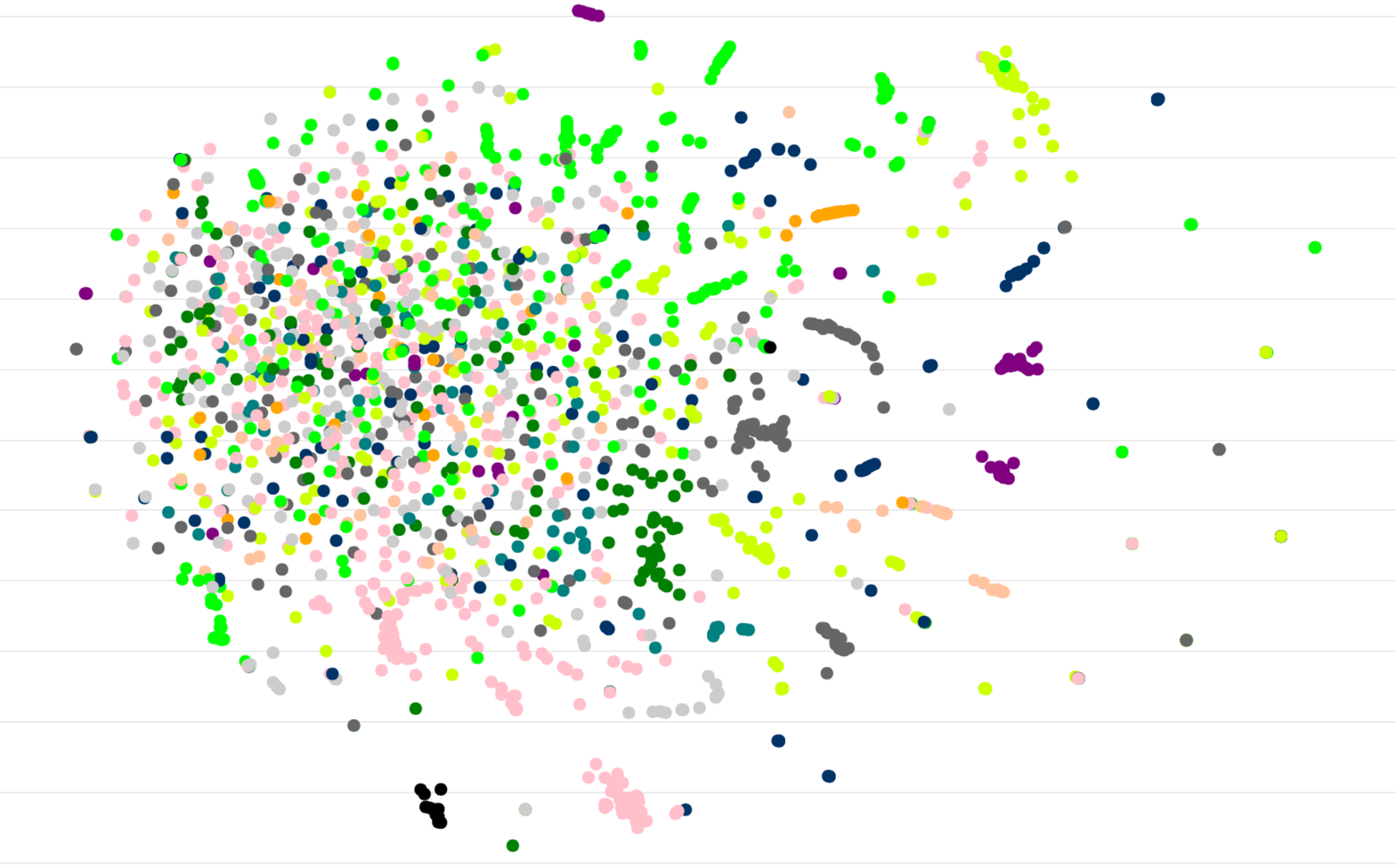}
  \vspace*{-2mm}
  \caption{Scatterplot of the final representations $\gb_i$'s of RandomDAG}
  \label{fig:tsne_alg_fake}
\end{subfigure}

\begin{subfigure}{.5\textwidth}
  \centering\captionsetup{width=.9\linewidth}%
  \includegraphics[width=.55\linewidth]{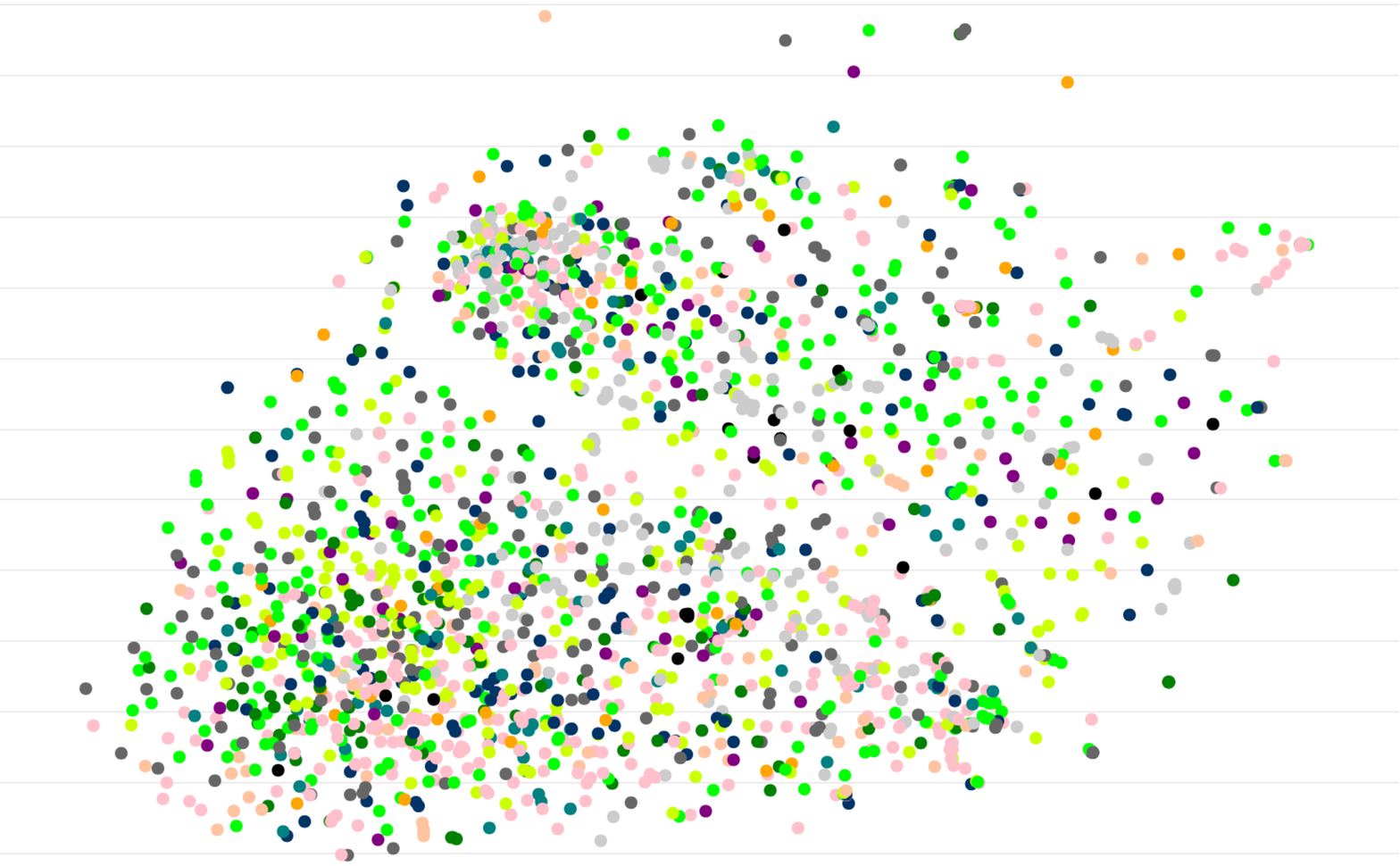}
  \vspace*{-2mm}
  \caption{Scatterplot of the disease representations trained by GloVe}
  \label{fig:tsne_glove}
\end{subfigure}%
\begin{subfigure}{.5\textwidth}
  \centering\captionsetup{width=.9\linewidth}%
  \includegraphics[width=.55\linewidth]{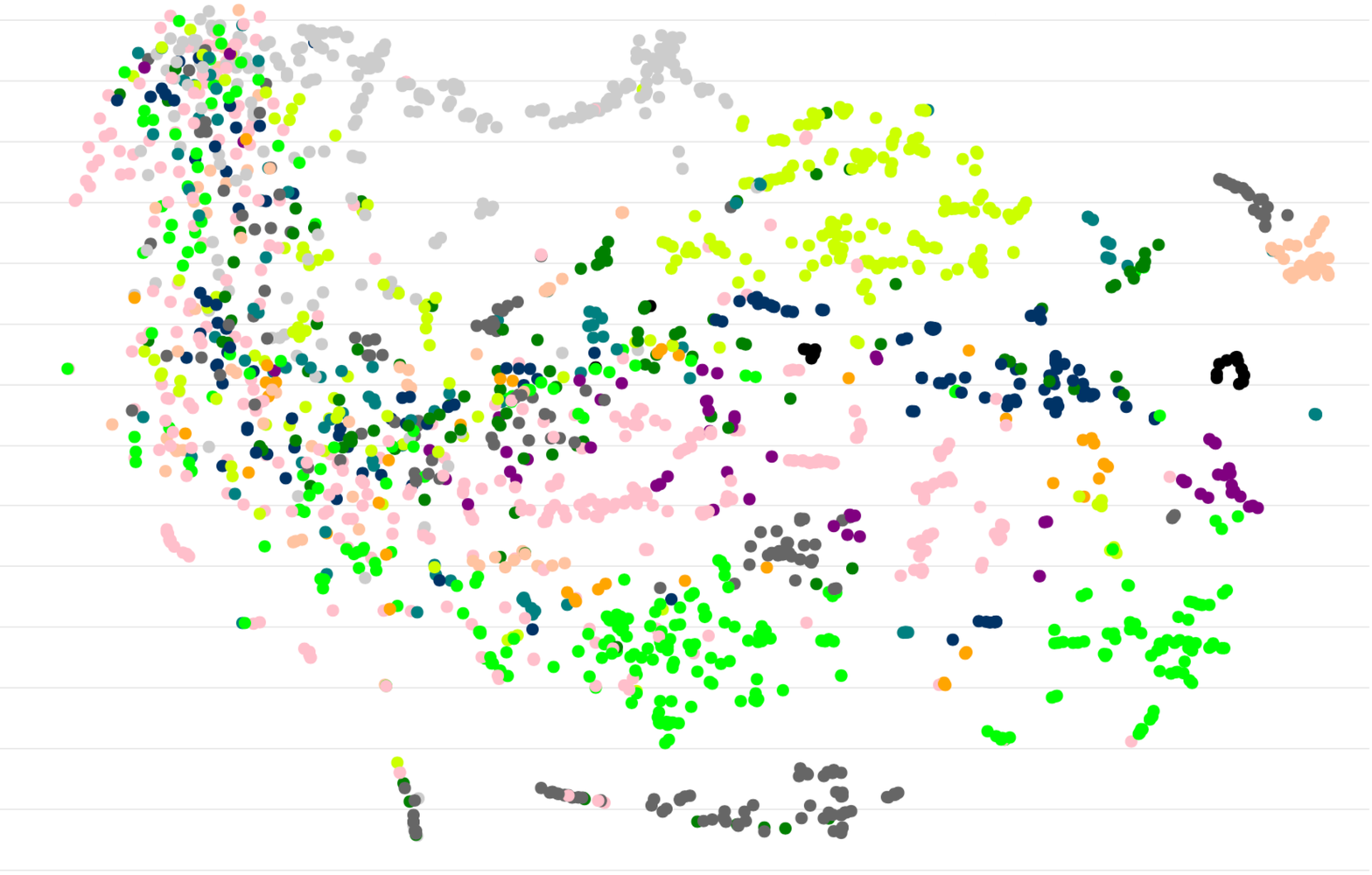}
  \vspace*{-2mm}
  \caption{Scatterplot of the basic embeddings $\eb_i$'s trained by Skip-gram}
  \label{fig:tsne_skipgram}
\end{subfigure}
\vspace*{-2mm}
\caption{t-SNE scatterplots of medical concepts trained by \texttt{GRAM}+, \texttt{GRAM}, RNN+, RNN, RandomDAG, GloVe and Skip-gram. The color of the dots represents the highest disease categories and the text annotations represent the detailed disease categories in CCS multi-level hierarchy. It is clear that \texttt{GRAM}+ and \texttt{GRAM} exhibit interpretable embedding that are well aligned with the medical ontology.}
\label{fig:tsne_main}
\vspace*{-8mm}
\end{figure*}

To qualitatively assess the interpretability of the learned representations of the medical codes, we plot on a 2-D space using t-SNE \citep{maaten2008visualizing} the final representations $\gb_i$ of 2,000 randomly chosen diseases learned by \texttt{GRAM}+ for sequential diagnoses prediction on Sutter data\footnote{The scatterplots of models trained for sequential diagnoses prediction on MIMIC-III and HF prediction for Sutter HF cohort were similar but less structured due to smaller data size.} (Figure \ref{fig:tsne_alg_glove}).
The color of the dots represents the highest disease categories and the text annotations represent the detailed disease categories in CCS multi-level hierarchy. 
For comparison, we also show the t-SNE plots on the strongest results from \mname (Figure \ref{fig:tsne_alg}), RNN+ (Figure \ref{fig:tsne_rnn_glove}), RNN (Figure \ref{fig:tsne_rnn}) and RandomDAG (Figure \ref{fig:tsne_alg_fake}). GloVe (Figure \ref{fig:tsne_glove}) and Skip-gram (Figure \ref{fig:tsne_skipgram}) were trained on the Sutter data, where a single visit $V_t$ was used as the context window to calculate the co-occurrence of codes. 

Figures \ref{fig:tsne_rnn_glove} and \ref{fig:tsne_glove} confirm that interpretable representations cannot simply be learned only by co-occurrence or supervised prediction without medical knowledge.
\texttt{GRAM}+ and \texttt{GRAM} learn interpretable disease
\newpage
\clearpage
\newpage
\clearpage
representations that are significantly more consistent with the given knowledge DAG $\mathcal{G}$. Based on the prediction performance shown by Table \ref{tab:predictions}, and the fact that the representations $\gb_i$'s are the final product of \texttt{GRAM}, we can infer that such medically meaningful representations are necessary for predictive models to cope with data insufficiency and make more accurate predictions.
Figure \ref{fig:tsne_alg} shows that the quality of the final representations $\gb_i$ of \mname is quite similar to \texttt{GRAM}+. Compared to other baselines, \mname demonstrates significantly more structured representations that align well with the given knowledge DAG. It is interesting that Skip-gram shows the most structured representation among all baselines. We used GloVe to initialize the basic embeddings $\eb_i$ in this work because it uses global co-occurrence information and its training time is fast as it is only dependent only on the total number of unique concepts $|\mathcal{C}|$. Skip-gram's training time, on the other hand, depends on both the number of patients and the number of visits each patient made, which makes the algorithm generally slower than GloVe.
An interactive visualization tool can be accessed at \url{http://www.sunlab.org/research/gram-graph-based-attention-model/}.

%\vspace*{-2mm}
\subsection{Analysis of the attention behavior}
\label{ssec:exp_attention}
\begin{figure*}[h]
\centering
\includegraphics[width=1.0\textwidth]{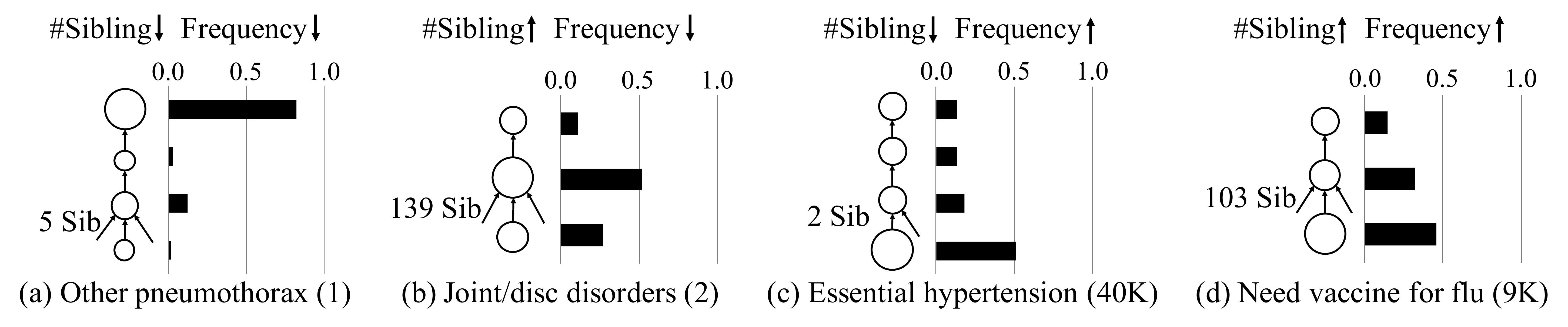}
\caption{\mname's attention behavior during HF prediction for four representative diseases (each column). In each figure, the leaf node represents the disease and upper nodes are its ancestors. The size of the node shows the amount of attention it receives, which is also shown by the bar charts. The number in the parenthesis next to the disease is its frequency in the training data. We exclude the root of the knowledge DAG $\mathcal{G}$ from all figures as it did not play a significant role.}
\label{fig:attentions}
\vspace{-0.1in}
\end{figure*}
Next we show that \mname's attention can be explained intuitively based on the data availability and knowledge DAG's structure when performing a prediction task. %\jsedit{The examples all seem not directly related to the target. I wonder if the attention weights will vary across tasks. more specifically, how/if attention weights across concepts will be different when we predict HF vs Cancer. When we predict HF, will all the leaf cardiovascular concepts have higher weights than predicitng cancer?}
Using Eq. \eqref{eq:final_representation}, we can calculate the attention weights of individual disease. Figure \ref{fig:attentions} shows the attention behaviors of four representative diseases when performing HF prediction on Sutter HF cohort.

\textit{Other pneumothorax} (ICD9 512.89) in Figure \ref{fig:attentions}a is rarely observed in the data and has only five siblings. In this case, most information is derived from the highest ancestor. 
\textit{Temporomandibular joint disorders \& articular disc disorder} (ICD9 524.63) in Figure \ref{fig:attentions}b is rarely observed but has 139 siblings. In this case, its parent receives a stronger attention because it aggregates sufficient samples from all of its children to learn a more accurate representation. Note that the disease itself also receives a stronger attention to facilitate easier distinction from its large number of siblings.

\textit{Unspecified essential hypertension} (ICD9 401.9) in Figure \ref{fig:attentions}c is very frequently observed but has only two siblings. In this case, \mname assigns a very strong attention to the leaf, which is logical because the more you observe a disease, the stronger your confidence becomes. \textit{Need for prophylactic vaccination and inoculation against influenza} (ICD9 V04.81) in Figure \ref{fig:attentions}d is quite frequently observed and also has 103 siblings. The attention behavior in this case is quite similar to the case with fewer siblings (Figure \ref{fig:attentions}b) with a slight attention shift towards the leaf concept as more observations lead to higher confidence. 

\section{Related Work}
\label{sec:related}
The attention mechanism is a general framework for neural network learning \citep{bahdanau2014neural}, and has been since used in many areas such as speech recognition \citep{chorowski2014end}, computer vision \citep{ba2014multiple,xu2015show} and healthcare \citep{choi2016retain}. However, no one has designed attention model based on knowledge ontology, which is the focus of this work.

There are related works in learning the representations of graphs.  Several studies focused on learning the representations of graph vertices by using the neighbor information. DeepWalk \citep{Perozzi:2014:DOL:2623330.2623732} and node2vec \citep{Grover:2016:NSF:2939672.2939754} use random walk while LINE \citep{Tang:2015:LLI:2736277.2741093} uses breadth-first search to find the neighbors of a vertex and learn its representation based on the neighbor information. Graph convolutional approaches \citep{yang2016revisiting,kipf2016semi} also focus on learning the vertex representations to mainly perform vertex classification. All those works focus on solving the graph data problems whereas \mname focuses on solving clinical predictive modeling problems using the knowledge DAG as supplementary information.

Several researchers tried to model the knowledge DAG such as WordNet \citep{miller1995wordnet} or Freebase \citep{bollacker2008freebase} where two entities are connected with various types of relation, forming a set of triples. 
They aim to project entities and relations \citep{bordes2013translating, socher2013reasoning, wang2014knowledge, lin2015learning} to the latent space based on the triples or additional information such as hierarchy of entities \citep{xie2016hierarchical}. These works demonstrated tasks such as link prediction, triple classification or entity classification using the learned representations.
More recently, \citet{li2016joint} learned the representations of words and Wikipedia categories by utilizing the hierarchy of Wikipedia categories. %Their goal is to incorporate hierarchical information into the the learned representations, and they demonstrated word categorization and document classification using the learned representations.
\mname is fundamentally different from the above studies in that it aims to design intuitive attention mechanism on the knowledge DAG as a knowledge prior to cope with data insufficiency and learn medically interpretable representations to make accurate predictions.% Furthermore, we employ the attention mechanism on the knowledge DAG to dynamically combine the nodes, and the entire process is trained in a supervised fashion as opposed to unsupervised methods above. 

A classical approach for incorporating side information in the predictive models is to use graph Laplacian regularization \citep{weinberger2006graph,che2015deep}. 
%In our setting, this approach can be applied by first constructing a distance metric for the medical concepts based on the ontological graph. Then, we can use the distance metric in the graph Laplacian regularizer. 
However, using this approach is not straightforward as it relies on the appropriate definition of distance on graphs which is often unavailable.

%\vspace*{-2mm}
\section{Conclusion}
\label{sec:conclusion}
Data insufficiency, either due to less common diseases or small datasets, is one of the key hurdles in healthcare analytics, especially when we apply deep neural networks models. To overcome this challenge, we leverage the knowledge DAG, which provides a multi-resolution view of medical concepts. We propose \mname, a graph-based attention model using both a knowledge DAG and EHR to learn an accurate and interpretable representations for  medical concepts. \mname chooses a weighted average of ancestors of a medical concept and train the entire process with a predictive model in an end-to-end fashion. We conducted three predictive modeling experiments on real EHR datasets and showed significant improvement in the prediction performance, especially on low-frequency diseases and small datasets. Analysis of the attention behavior provided intuitive insight of \mname. 
%\ecedit{For future work, we plan to incorporate medication and procedure information into \mname framework by using an appropriate medication hierarchy and a procedure hierarchy. We expect this to improve the prediction performance even further.}

%\section{Discussion}
%\label{sec:discuss}
%\input{discuss.tex}

\bibliographystyle{ACM-Reference-Format}
%\vspace*{-3mm}
\bibliography{references}
\end{document}